\documentclass[letterpaper, 10 pt, journal, twoside]{IEEEtran}

\usepackage{adjustbox}
\usepackage{amsmath}
\usepackage{amssymb}
\usepackage{booktabs}
\usepackage[style=base]{caption}
\usepackage{comment}
\usepackage{gensymb}
\usepackage{graphicx}
\usepackage{mathtools}
\usepackage{multirow}
\usepackage{multicol}
\usepackage[percent]{overpic}
\usepackage{pifont}
\usepackage{xspace}
\usepackage{xfrac}
\usepackage{wrapfig,lipsum}

\usepackage[table]{xcolor}
\definecolor{lightgray}{RGB}{131, 126 114}
\definecolor{darkgreen}{RGB}{0, 150, 0}
\definecolor{darkred}{RGB}{200, 0, 0}
\definecolor{darkblue}{RGB}{0, 0, 200}
\definecolor{gray}{RGB}{142,142,142}

\newcommand{\lone}{$L_1$\xspace}
\makeatletter
\DeclareRobustCommand\onedot{\futurelet\@let@token\@onedot}
\def\@onedot{\ifx\@let@token.\else.\null\fi\xspace}
\def\eg{\emph{e.g}\onedot} 
\def\ie{\emph{i.e}\onedot} 
\def\etal{\emph{et al}\onedot}
\makeatother
\newcommand{\ch}{{\color{darkgreen} \ding{51}}}
\newcommand{\xm}{{\color{darkred} \ding{55}}}
\newcommand{\xmb}{{\color{darkblue} \ding{55}}}
\newcommand{\nap}{{\color{gray} NA}}

\usepackage[colorlinks,pagebackref=false,citecolor=blue,bookmarks=false,hypertexnames=true]{hyperref}
\begin{document}
\title{OmniDet: Surround View Cameras based Multi-task Visual Perception Network for Autonomous Driving}

\author{Varun Ravi Kumar$^{1,5\dag}$, Senthil Yogamani$^{2\dag}$, Hazem Rashed$^{3}$, Ganesh Sistu$^{2}$, \\
Christian Witt$^{1}$, Isabelle Leang$^{4}$, Stefan Milz$^{5}$ and Patrick M\"ader$^{5}$ \\
{
$^{1}$Valeo, Germany\quad
$^{2}$Valeo, Ireland\quad
$^{3}$Valeo, Egypt\quad
$^{4}$Valeo, France\quad
$^{5}$TU Ilmenau, Germany\quad
$^\dag$co-first authors
}

\thanks{Manuscript received: Oct, 15\textsuperscript{th}, 2020; 
Revised Jan, 9\textsuperscript{th}, 2021; 
Accepted Feb, 3\textsuperscript{rd}, 2021. 
This paper was recommended for publication by Editor Eric Marchand upon evaluation of the Associate Editor and Reviewers' comments.}

\thanks{Digital Object Identifier (DOI): see top of this page.}}


\maketitle
\vspace*{-1cm}
\begin{abstract}
Surround View fisheye cameras are commonly deployed in automated driving for 360\degree{} near-field sensing around the vehicle. This work presents a multi-task visual perception network on unrectified fisheye images to enable the vehicle to sense its surrounding environment. It consists of six primary tasks necessary for an autonomous driving system: depth estimation, visual odometry, semantic segmentation, motion segmentation, object detection, and lens soiling detection. We demonstrate that the jointly trained model performs better than the respective single task versions. Our multi-task model has a shared encoder providing a significant computational advantage and has synergized decoders where tasks support each other. We propose a novel camera geometry based adaptation mechanism to encode the fisheye distortion model both at training and inference. This was crucial to enable training on the WoodScape dataset, comprised of data from different parts of the world collected by 12 different cameras mounted on three different cars with different intrinsics and viewpoints. 
Given that bounding boxes is not a good representation for distorted fisheye images, we also extend object detection to use a polygon with non-uniformly sampled vertices. We additionally evaluate our model on standard automotive datasets, namely KITTI and Cityscapes. We obtain the state-of-the-art results on KITTI for depth estimation and pose estimation tasks and competitive performance on the other tasks. We perform extensive ablation studies on various architecture choices and task weighting methodologies. A short video at {\url{https://youtu.be/xbSjZ5OfPes}} provides qualitative results.
\end{abstract}
\section{Introduction}

Surround View fisheye cameras have been deployed in premium cars for over ten years, starting from visualization applications on dashboard display units to providing near-field perception for automated parking. Fisheye cameras have a strong radial distortion that cannot be corrected without disadvantages, including reduced field-of-view and resampling distortion artifacts at the periphery \cite{kumar2020unrectdepthnet}. Appearance variations of objects are larger due to the spatially variant distortion, particularly for close-by objects. Thus fisheye perception is a challenging task, however it is relatively less explored despite its prevalence.\par
Autonomous Driving applications require various perception tasks to provide a robust system covering a wide variety of use cases. Alternate ways to detect objects in parallel are necessary to achieve a high level of accuracy. For example, objects can be detected based on appearance, motion, and depth cues. Despite increasing computation power in automotive embedded systems, there is always a need for efficient design due to the increasing number of cameras and perception tasks. Multi-task learning (MTL) is an efficient design pattern commonly used where most of the computation is shared across all the tasks~\cite{sistu2019neurall, vandenhende2020multitask}. Furthermore, learning features for multiple tasks can act as a regularizer, improving generalization. Mao~\etal~\cite{mao2020multitask} illustrated that multi-task learning improves adversarial robustness, which is critical for safety applications. In the automotive multi-task setting, MultiNet~\cite{teichmann2018multinet} was one of the first to demonstrate a three task network on KITTI, and further works have primarily worked on a three task setting.\par
\begin{figure}[!t]
  \vspace{-0.5cm}
  \captionsetup{singlelinecheck=false, font=footnotesize, skip=2pt, belowskip=-14pt}
  \resizebox{\columnwidth}{!}{
  \centering
\begin{tabular}{ccc}
 	\begin{overpic}[width=0.7\columnwidth]{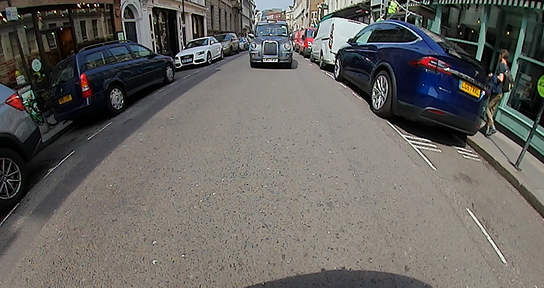}
    \put (0,3) {\colorbox{lightgray}{$\displaystyle\textcolor{black}{\text{(a)}}$}}
    \end{overpic}
    \begin{overpic}[width=0.7\columnwidth]{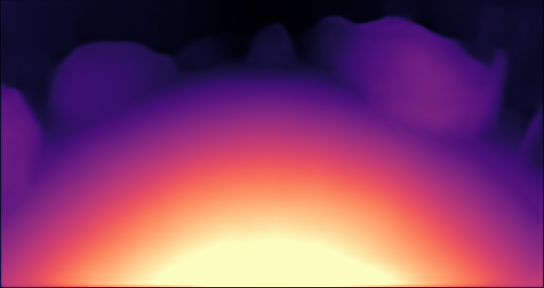}
    \put (0,3) {\colorbox{lightgray}{$\displaystyle\textcolor{black}{\text{(b)}}$}}
    \end{overpic} \\
    
    \begin{overpic}[width=0.7\columnwidth]{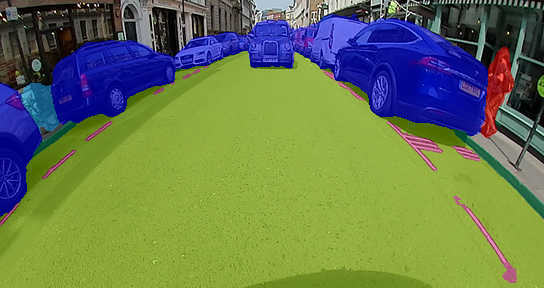}
    \put (0,3) {\colorbox{lightgray}{$\displaystyle\textcolor{black}{\text{(c)}}$}}
    \end{overpic} 
 	\begin{overpic}[width=0.7\columnwidth]{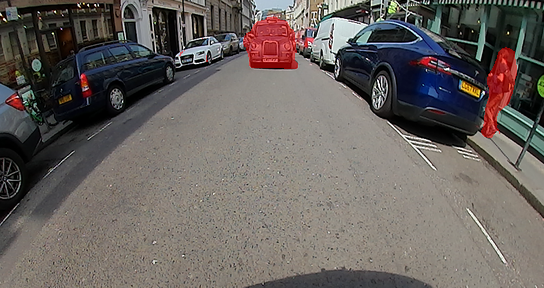}
    \put (0,3) {\colorbox{lightgray}{$\displaystyle\textcolor{black}{\text{(d)}}$}}
    \end{overpic} \\
    
    \begin{overpic}[width=0.7\columnwidth]{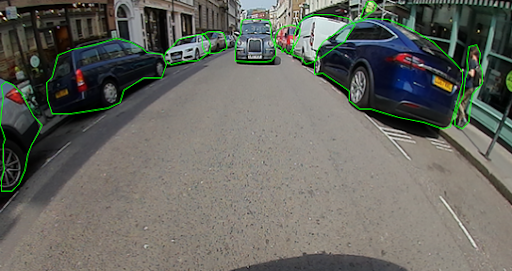}
    \put (0,3) {\colorbox{lightgray}{$\displaystyle\textcolor{black}{\text{(e)}}$}}
    \end{overpic}
    \begin{overpic}[width=0.7\columnwidth]{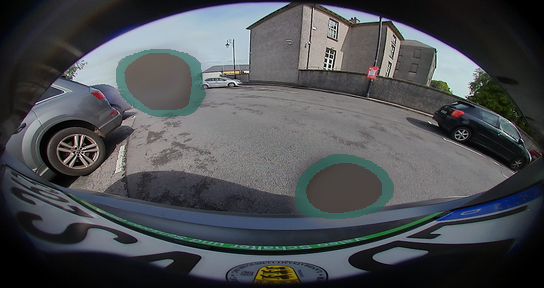}
    \put (0,3) {\colorbox{lightgray}{$\displaystyle\textcolor{black}{\text{(f)}}$}}
    \end{overpic}
\end{tabular}
}
\caption{\textbf{Results of our real-time OmniDet model on raw fisheye images.} (a) Rear-Camera Input Image, (b) Distance Estimation, (c) Semantic Segmentation, (d) Motion Estimation, (e) 24-sided Polygon based Object Detection and (f) Soiling Segmentation (asynchronous).}
\label{fig:abstract}
\end{figure}
This work demonstrates a multi-task perception model for the six essential perception tasks on unrectified fisheye images (shown in Fig. \ref{fig:abstract}). As we discuss a full perception system building upon a lot of our previous work, it is difficult to cover all the technical details, and we have focused on the main contributions. Our contributions are as follows:
\begin{itemize}
    \item We demonstrate the first real-time six-task model for surround-view fisheye camera perception.
    \item We propose a novel camera tensor representation of radial distortion to enable the adaptation of CNN for the 12 different camera models in the WoodScape dataset.
    \item We propose novel design techniques, including VarNorm task weighting.
    \item We design synergized decoders where different tasks help each other in addition to a shared encoder.
    \item We showcase a six-task model on WoodScape and five-task model on KITTI and Cityscapes performing better than the single task baselines.
    \item We obtain state-of-the-art results for depth and pose estimation tasks on KITTI among monocular methods.
\end{itemize}
\section{Perception tasks and Losses in our MTL}

Our goal is to build a multi-task model covering the necessary modules for near-field sensing use cases like Parking or Traffic Jam assist. This paper builds upon our previous papers focused on individual tasks, and we mainly discuss the new improvements. In the following sub-sections, we refer the reader to these papers for more details and a literature review. In general, there is minimal work in the area of fisheye perception. Specifically, there is only our previous work on multi-task learning: FisheyeMultiNet \cite{maddu2019fisheyemultinet} which discusses a more straight forward three task network.\par

The perception system comprises semantic tasks, geometric tasks, and lens soiling detection. The standard semantic tasks are object detection (pedestrians, vehicles, and cyclists) and semantic segmentation (road, lanes, and curbs).
Fisheye cameras are mounted low on a vehicle and are susceptible to lens soiling due to splashing of mud or water from the road. Thus, it is vital to detect soiling on the camera lens and trigger a cleaning system~\cite{uvrivcavr2019soilingnet}.
The semantic tasks typically require a large annotated dataset covering various objects. It is infeasible to practically cover every possible object.
Thus, generic object detection using geometric cues like motion or depth for rare objects is typically used. They will also complement the detection of standard objects and provide more robustness.
Thus we propose to include motion segmentation and depth estimation tasks. Motion is a dominant cue in automotive scenes, and it requires at least two frames or the use of dense optical flow~\cite{siam2018modnet}. Self-supervised methods have recently dominated depth estimation, which has also been demonstrated on fisheye images~\cite{kumar2020fisheyedistancenet}.
Finally, the visual odometry task is required to place the detected objects in a temporally consistent map.\par
\subsection{Self-Supervised Distance and Pose Estimation Networks}

We set up a self-supervised monocular structure-from-motion (SfM) framework using our previous work FisheyeDistanceNet~\cite{kumar2020fisheyedistancenet} for distance and pose estimation and perform view synthesis by incorporating the polynomial projection model function. The total loss comprised of a reconstruction matching term $\mathcal{L}_r$, regularization term $\mathcal{L}_s$, which enforces edge-aware smoothness inside the distance map $\hat{D}_t$ as introduced in~\cite{Godard2019}. Additionally, the cross-sequence distance consistency loss $\mathcal{L}_{dc}$ and the scale recovery technique were used. We discuss the new improvements in the following paragraphs, which led to significant improvement in accuracy.\par

We incorporate feature-metric losses from~\cite{shu2020featdepth} where $\mathcal{L}_{dis}$ and $\mathcal{L}_{cvt}$ are computed on $I_t$'s feature representation, where we learn the features using a self-attention autoencoder. The main goal of these losses is to prevent the training objective from getting stuck at multiple local minima for homogeneous areas as fisheye images have considerably larger homogeneous areas than their rectilinear counterpart. It is essentially a loss function that penalizes small slopes and emphasizes the low-texture regions using the image gradients. The self-supervised loss landscapes are constrained to form proper convergence basins using the first-order derivatives to regularize the target features. However, merely imposing a discriminative loss cannot guarantee we move to the optimal solution during the gradient descent, since inconsistency exists among first-order gradients, \ie spatially adjacent gradients point in opposite directions. Shu~\etal~\cite{shu2020featdepth} proposed a convergent loss to have a relatively large convergence radius to enable gradient descent from a remote distance. This is achieved by formulating the loss to have consistent gradients during the optimization step by encouraging the smoothness of feature gradients and large convergence radii accordingly. The total objective loss for distance estimation $\mathcal{L}_{dist}$ is
\begin{align}
    \mathcal{L}_{dist} &= \mathcal{L}_r(I_t,\hat{I}_{t'\to t}) + \beta~\mathcal{L}_s(\hat{D}_t) + \gamma~\mathcal{L}_{dc}(\hat{D}_t,\hat{D}_{t'}) \\
    &+ \mathcal{L}_r(\hat{F}_t,\hat{F}_{t'\to t}) + \omega~\mathcal{L}_{dis}(I_t, \hat{F}_t) 
    + \mu~\mathcal{L}_{cvt}(I_t, \hat{F}_t) \nonumber
    \label{eq:overall-loss}
\end{align}
where $\beta$, $\gamma$, $\omega$ and $\mu$ weigh the distance regularization $\mathcal{L}_s$, cross-sequence distance consistency $\mathcal{L}_{dc}$, discriminative $\mathcal{L}_{dis}$ and convergent $\mathcal{L}_{cvt}$ losses respectively.\par
We calculate the image and feature reconstruction loss using the target $I_t$, estimated feature $\hat{F}_t$ frames, reconstructed target $\hat I_{t' \to t}$ and feature $\hat{F}_{t'\to t}$ frames. It is a linear combination of the general robust pixel-wise loss term~\cite{barron2019general} and the Structural Similarity (SSIM)~\cite{wang2004image} as described in~\cite{kumar2020syndistnet}.\par
\subsection{Generalized Object Detection}

The standard bounding box representation fails in fisheye cameras due to heavy radial distortion, particularly in the periphery. In concurrent work~\cite{rashed2021generalized}, we explored different output representations for fisheye images, including oriented bounding boxes, curved boxes, ellipses, and polygons. We have integrated this model in our MTL framework where we use a 24-sided polygon representation for object detection. We briefly summarize the details here and refer to our expanded paper~\cite{rashed2021generalized} for more details on generalized object detection.\par

A polygon is a generic representation for any arbitrary shape; however, it is more expensive to annotate than a bounding box. The object contour can be uniformly sampled in the range of $360\degree$ split into $N$ equal polygon vertices, each represented by the radial distance $r$ from the object's centroid as used in PolyYOLO~\cite{polyyolo}. We observe that uniform sampling cannot efficiently represent high curvature variations in the fisheye image object contours. Thus, we make use of an adaptive sampling based on the curvature of the local contour. We distribute the vertices non-uniformly in order to optimally represent the object contour. We adopt the algorithm in~\cite{teh1989detection} to detect the dominant points in a given curved shape, which best represents the object. Then we reduce the set of points using the algorithm in~\cite{douglas1973algorithms} to get the most representative simplified curves. We adapted the YOLOv3~\cite{YOLOV3} decoder to output polygons and other representations listed above for a uniform comparison.\par
\vspace{-3pt}
\subsection{Segmentation Tasks}
\label{sec:segmentation-tasks}

Three of our tasks are modeled as segmentation problems. Semantic and soiling segmentation having seven and four output classes, respectively, on the WoodScape dataset. Motion segmentation uses two frames and outputs a binary moving or static mask. During training, the network predicts the posterior probability $Y_t$ which is optimized in a supervised fashion by \textit{Lovasz-Softmax}~\cite{berman2018lovasz} loss, and \textit{Focal}~\cite{lin2017focal} loss for handling class imbalance instead of the cross-entropy loss used in our previous work. We obtain the final segmentation mask $M_{t}$ by applying a pixel-wise $\operatorname{argmax}$ operation on the posterior probabilities.\par

The soiling dataset is independently built, and thus it cannot be trained jointly in a traditional manner. Thus we freeze the shared encoder trained using five other tasks and train only the decoder for soiling. This demonstrates the potential of re-using the encoder features for additional tasks. We also trained soiling jointly using asynchronous backpropagation \cite{kokkinos2017ubernet}, but it achieved the same accuracy as using the frozen encoder. Compared to our previous work SoilingNet \cite{uvrivcavr2019soilingnet}, we moved from the tiled output to pixel-level segmentation.\par
\begin{figure*}[t]
  \captionsetup{singlelinecheck=false, font=footnotesize, skip=2pt, belowskip=-8pt}
  \centering
    \includegraphics[width=\textwidth]{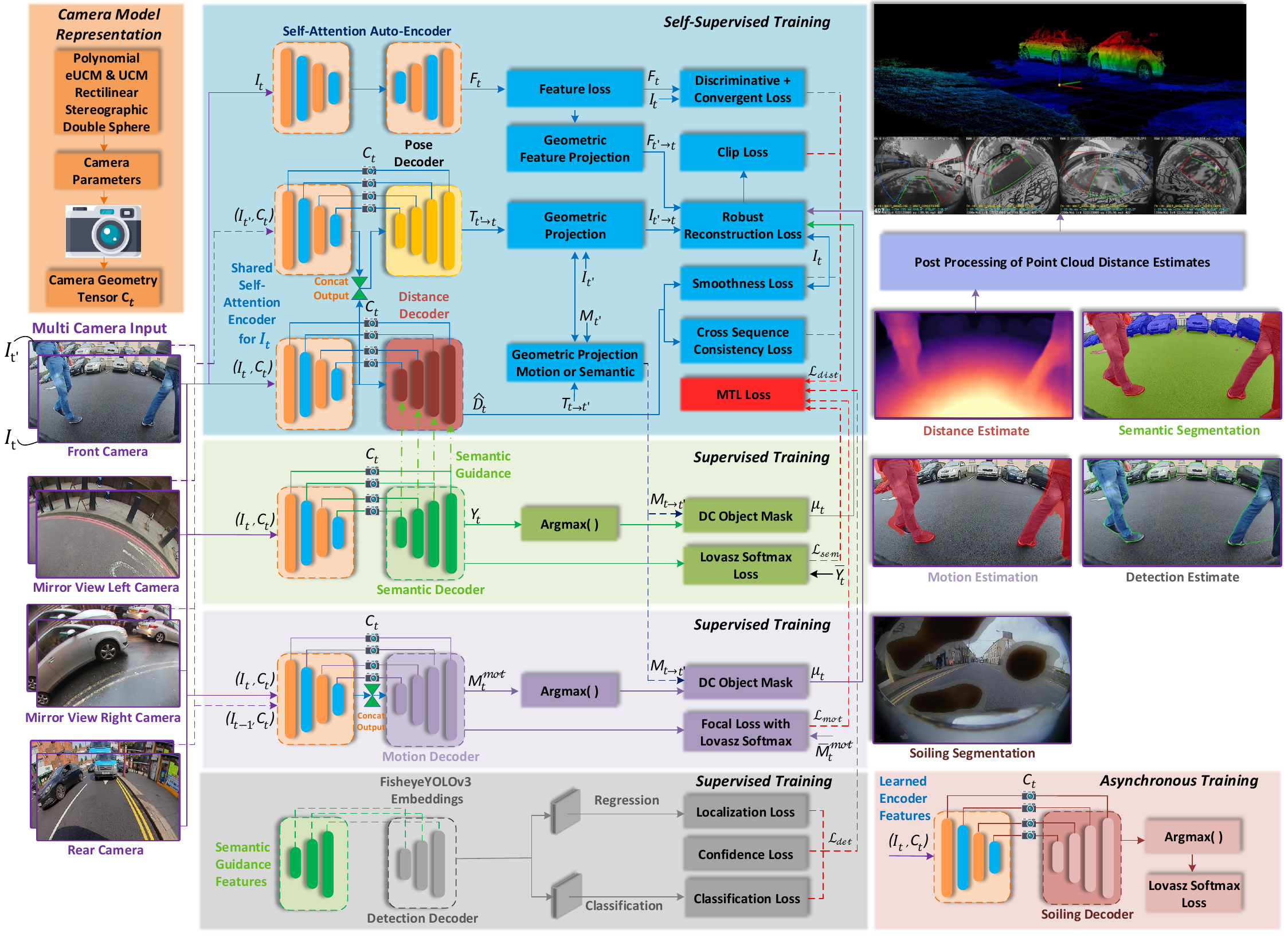}
    \caption{\textbf{Overview of our Surround View cameras based multi-task visual perception framework.} The distance estimation task (\textcolor[HTML]{00b0f0}{blue} block) makes use of semantic guidance and dynamic object masking from semantic/motion estimation (\textcolor[HTML]{00b050}{green} and \textcolor[HTML]{ab9ac0}{blue haze} block) and camera-geometry adaptive convolutions (\textcolor{orange}{orange} block). Additionally, we guide the detection decoder features (\textcolor{gray}{gray} block) with the semantic features. The encoder block (shown in the same color) is common for all the tasks.
    Our framework consists of processing blocks to train the self-supervised distance estimation (\textcolor[HTML]{00b0f0}{blue} blocks) and semantic segmentation (\textcolor[HTML]{00b050}{green} blocks), motion segmentation (\textcolor[HTML]{ab9ac0}{blue haze} blocks), polygon-based fisheye object detection (\textcolor{gray}{gray} blocks), and the asynchronous task of soiling segmentation (\textcolor[HTML]{d9958f}{rose fog} block). We obtain Surround View geometric information by post-processing the predicted distance maps in 3D space (\textcolor[HTML]{a7b4ef}{perano} block). The camera tensor $C_t$ (\textcolor{orange}{orange} block) helps our OmniDet yield distance maps on multiple camera-viewpoints and make the network camera independent. 
    }
    \label{fig:mtl_pipeline}
\end{figure*}
\vspace{-4pt}
\subsection{Joint Optimization}

Balancing the task losses is an important problem for training multi-task models. Our contribution is two-fold. We evaluate various task weighting strategies for five tasks compared to the three tasks experiments in previous literature. We evaluate the uncertainty loss from Kendall~\cite{Kendall2018}, the gradient magnitude normalization GradNorm~\cite{chen2018gradnorm}, the dynamic task prioritization DTP~\cite{guo2018dynamic}, the dynamic weight average DWA~\cite{liu2019end} and the geometric loss~\cite{chennupati2019multinet++}.\par

Secondly, we propose a novel method called \textbf{VarNorm} for variance normalization. It consists of normalizing each loss by its variance over the last $n$ epochs. The loss weight of task $i$ at epoch $t$ is formulated as below:
\begin{equation}
    w_i(t) = \frac{1}{\sigma_i(t-1)},
    \sigma_i(t) = \frac{1}{n-1} \sum_{k=0}^{n-1} (L_i(t-k) - \overline{L_i})^2
    \label{eq:varnorm}
\end{equation}
where $\overline{L_i}$ is the average of task loss $i$ over the last $n$ epochs. We chose $n=5$. This method is motivated by the simple idea that the task loss values can be seen as a distribution whose dispersion is its variance. Variance normalization re-scales the dispersion between the different task loss distributions based on the previous $n$ epochs. A large dispersion leads to a lower task weight, whereas a small dispersion to a higher one. Its final effect tends to homogenize the learning speed of tasks. As shown in Table~\ref{table:task-weighting}, equal weighting is the worst, and our multi-task network performs better than the single task networks by using any dynamic task weighting method presented above. We employ the proposed VarNorm method for all the further experiments as it achieved the best results.\par
\section{Network Details of Mtl OmniDet} \label{sec:network-details}

Encoder-decoder architectures are commonly used for dense prediction tasks. We use this type of architecture as it easily extends to a shared encoder for multiple tasks. We design our encoder by incorporating vector attention based pairwise and patchwise self-attention encoders from~\cite{zhao2020exploring}. These networks efficiently adapt the weights across both spatial dimensions and channels. We adapt the Siamese (twin network) approach for the motion prediction network, where we concatenate the source and target frame features and pass them to the super-resolution motion decoder. As the weights are shared in the Siamese encoder, the previous frame's encoder can be saved and re-used instead of re-computing. Inspired by~\cite{shu2020featdepth}, we develop an auxiliary self-attention auto-encoder for single-view reconstruction.
We detail our main novel contributions in the next two sub-sections. Firstly, we employ our novel camera geometry tensor to handle multiple viewpoints and change in the camera's intrinsic distance estimation. Secondly, we employ synergized decoders via cross task connections to improve each other's performance.\par
\subsection{Camera Geometry Tensor $C_t$}
\label{sec:camera-geometry-tensor}

\subsubsection{Motivation} 

In an industrial deployment setting, we target the design of a model that can be deployed in millions of vehicles each having its own set of cameras. Although the underlying camera intrinsics model is the same for a particular family of vehicles, there are variations due to manufacturing processes, which require the calibration of each camera instance. Even after deployment, calibration can vary due to high environmental temperature or due to aging. Thus a calibration adaptation mechanism in the model is essential. This contrasts with public datasets, which have a single camera instance for both the training and test dataset. In the Woodscape fisheye dataset~\cite{yogamani2019woodscape}, there are 12 different cameras with slight intrinsic variations to evaluate this effect. A single model for these four cameras instead of four individual models would also have several practical advantages such as (1) an improved efficiency on the embedded system requiring less memory and data rate to transmit, (2) an improved training by access to a larger dataset and regularization through different views, and (3) maintenance and certification of one model instead of four.\par 

We propose converting all camera geometry properties into a tensor called camera geometry tensor $C_t$ that is then passed to the CNN model to tackle this problem. The closest work is CAM-Convs~\cite{Facil2019}, which uses camera-aware convolutions for pinhole cameras. We build upon this work and generalize to arbitrary camera geometries, including fisheye cameras.\par
\subsubsection{Approach}

We introduce the camera geometry tensor $C_t$ in the mapping from RGB features to 3D information for the Self-Attention Network (SAN) encoder module, as shown in Fig.~\ref{fig:mtl_pipeline}. It is included in each self-attention stage and also applied to every skip-connection.
The camera geometry tensor $C_t$ is formulated in a three-step process:
For efficient training, the pixel-wise coordinates and angle of incidence maps are pre-computed. The normalized coordinates per pixel are used for these channels by incorporating information from the camera calibration. We concatenate these tensors and represent them by $C_t$ and pass it along with the input features to our SAN \emph{pairwise} and \emph{patchwise} operation modules. It comprises six channels in addition to the existing decoder channel inputs. The proposed approach can, in principle, be applied to any fisheye projection model of choice explained in~\cite{kumar2020unrectdepthnet}. The different maps included in our shared self-attention encoder are computed using the camera intrinsic parameters, where the distortion coefficients $a_1, a_2, a_3, a_4$ are used to create the angle of incidence maps $(a_x,a_y)$, $c_x, c_y$ are used to compute the principal point coordinate maps $(cc_x,cc_y)$ and the camera's sensor dimensions (width $w$ and height $h$) are utilized to formulate the normalized coordinate maps.\par
\subsubsection{Centered Coordinates ($cc$)}

Principal point position information is fed into the SAN's \emph{pairwise} and \emph{patchwise} operation modules by including the $cc_x$ and $cc_y$ coordinate channels centered at $(0,0)$. We concatenate $cc_x$ and $cc_y$ by resizing them using bilinear interpolation to match the input feature size. We formulate $cc_x$ and $cc_y$ channels as:
\begin{equation}
\label{eq:ccx}
    \resizebox{0.85\hsize}{!}{%
    $
    cc_x=\begin{pmatrix}
           \smash{0 -c_x}\\
           \smash{1 -c_x}\\
           \smash{\vdots} \\
           \smash{w-c_x}
    \end{pmatrix}\cdot
    \begin{pmatrix}
           1 \\
           1 \\
           \smash{\vdots} \\
           1
    \end{pmatrix}_{(h+1)\times 1}^{\intercal}
,
    cc_y=\begin{pmatrix}
           1 \\
           1 \\
           \smash{\vdots} \\
           1
    \end{pmatrix}_{(w+1)\times 1}\cdot\begin{pmatrix}
           \smash{0 -c_y}\\
           \smash{1 -c_y}\\
           \smash{\vdots} \\
           \smash{h - c_y}
    \end{pmatrix}^{\intercal}
    $
    }
\end{equation}
\subsubsection{Angle of Incidence Maps $(a_x,a_y)$}

For the pinhole (rectilinear) camera model, the horizontal and vertical angle of incidence maps are calculated from the $cc$ maps using the camera's focal length $f$: $ a_{ch}[i,j] = \arctan({cc_{ch}[i,j]}/{f})$,
where $ch$ is either $x$ or $y$ (refer to Eq.~\ref{eq:ccx}). For the different fisheye camera models, the angle of incidence maps can analogously be deduced by taking the inverse of the radial distortion functions $r(\theta)$ explained in~\cite{kumar2020unrectdepthnet}. Specifically, for the polynomial model used in this paper, the angle of incidence $\theta$ is formulated by calculating the fourth order polynomial roots of $r(\theta) = \sqrt{x_I^2 + y_I^2} = a_1 \theta + a_2 \theta^2 + a_3 \theta^3 + a_4 \theta^4$ through a numerical method. 
We store the pre-calculated roots in a lookup table for all pixel coordinates to achieve training efficiency and create the $a_x$ and $a_y$
maps by setting $x_I = cc_x[i,j], y_I = 0$ 
and $x_I = 0, y_I = cc_y[i,j]$ respectively.
\subsubsection{Normalized Coordinates ($nc$)} 

Additionally, we add two channels of normalized coordinates~\cite{liu2018intriguing, Facil2019} whose values vary linearly between $-1$ and $1$ with respect to the image coordinates. 
The channels are independent of the camera sensor's properties and characterize the spatial location in x and y directions. (\textit{\eg} a value of the $\hat{x}$ channel nearer to $1$ indicates that the feature is nearer to the right border of the image).
\vspace{-4pt}
\subsection{Synergized Tasks} \label{sec:synergy}

\subsubsection{Dealing With Dynamic Objects and Solving Infinite Depth Issue}
\label{sec:dynamic-object-mask}

As dynamic objects violate the static world assumption, information about their depth/distance is essential in autonomous driving; else, we would encounter the infinite depth issue during the inference stage and incur contamination to the reconstruction losses. We use the motion segmentation information to exclude potentially \textit{moving} dynamic objects, while the distance is learned from \textit{non-moving} dynamic objects. For this purpose, we define the pixel-wise mask $\mu_t$, which contains a $1$ if a pixel does not belong to a dynamic object from the current frame $I_t$ and also not to a wrongfully projected dynamic object from the reconstructed frames $\hat{I}_{t'\to t}$ and a $0$ otherwise. Accordingly, we predict a motion segmentation mask $M^{mot}_t$ belonging to the target frame $I_t$, as well as motion masks $M_{t'}$ for the source frames $I_{t'}$. Dynamic objects inside the source frame are canonically detected inside $M_t$. However, to obtain the wrongfully projected dynamic objects, we need to warp the motion masks by nearest-neighbor sampling to the target frame, yielding projected motion masks $M_{t' \to t}$. 
We propose an alternative technique to enable the filtering of the dynamic objects if the motion segmentation task is unavailable. We leverage the semantic segmentation output and follow a similar method as described above. By defining the set dynamic object classes $\mathcal{S}_{\mathrm{DC}} \subset \mathcal{S}$, we can then reduce the semantic segmentation mask to a binary mask, fulfilling the above conditions. Its elements at location $ij$:
\begin{equation}
\!\!\mu_{t, ij} = 
\!\left\{
\begin{array}{l}
1 ,\; M_{t, ij} \notin \mathcal{S}_{\mathrm{DC}}\; \land \; M_{t'\rightarrow t,ij} \notin \mathcal{S}_{\mathrm{DC}} \\
0,\; \mathrm{else} \\
\end{array}
\right.
\label{eq:semantic_mask}
\end{equation}
Dynamic objects can be masked through a pixel-wise multiplication of the mask with the reconstruction loss for images and features.\par
\subsubsection{Semantically-Guided Distance and Detection Decoder}
\label{sec:semantic-guidance}

Following our previous work~\cite{kumar2020syndistnet}, to better incorporate the semantic knowledge extracted from the segmentation branch of the multi-task network into the distance estimation, we incorporate it using pixel-adaptive convolutions~\cite{su2019pixel} (PAC) to distill the knowledge from the semantic features into the distance decoder. This, in particular, breaks up the spatial invariance of the convolutions and allows the incorporation of location-specific semantic knowledge into the multi-level distance features. As shown in Fig.~\ref{fig:mtl_pipeline} (green block), the features are extracted at different levels of the segmentation decoder. Here, an input signal $x$ to the pixel-adaptive convolution is processed as
\begin{equation}
x_{ij}' = \sum_{ab \in \mathcal{N}_k(ij)} K(F_{ij},F_{ab}) W [r_{a-i,b-j}]x_{ab} + B
\label{eq:pac}
\end{equation}
with pixel location $ij$, distance $r_{a-i,b-j}$ between pixel locations and a $k \times k$ neighbourhood window $\mathcal{N}_k(ij)$ around location $ij$. The elements of $x$ inside the neighbourhood window $\mathcal{N}_k(ij)$ are respectively used as input to the convolution with weights $W$, bias $B\in \mathbb{R}^1$ and a kernel function $K$, used to calculate the correlation between the semantic guidance features $F \in \mathbb{R}^D$, extracted from the segmentation branch.\par
\subsubsection{Linking Self-Attention and Semantic features to 2D detection}

To leverage the multi-task learning setup, at first, we extract the Self-Attention Network (SAN)~\cite{zhao2020exploring} encoder features and feed it as an input signal to the Eq.~\ref{eq:pac}. We bypass the spatial information from the SAN encoder to the semantic decoder and fuse these features (skip-connections). Finally, we fuse these features and the detection decoder embeddings by applying PAC and obtaining content-agnostic features. This novel fusion technique in our OmniDet framework significantly improves the detection decoder's accuracy, which can be seen in Table~\ref{table:features}.\par
\subsection{Implementation Details}

We use Pytorch and employ a single stage learning process for our OmniDet framework to facilitate network optimization. We incorporate the recently proposed SAN in our encoder. The authors proposed two convolution variants, namely \emph{pairwise} and \emph{patchwise}. We mainly use patchwise but perform an ablation study on pairwise. We employ the Ranger (RAdam~\cite{liu2019variance} + LookAhead~\cite{zhang2019lookahead}) optimizer to minimize the training objective function. We train the model for 20 epochs, with a batch size of 24 on a 24GB Titan RTX with an initial learning rate of ${4 \times {10}^{-4}}$ for the first 15 epochs, which is reduced to ${{10}^{-5}}$ for the last 5 epochs. The sigmoid output $\sigma$ from the distance decoder is converted to distance with $D = {m \cdot \sigma + n}$, where $m$ and $n$ are chosen to constrain $D$ between $0.1$ and $100$ units. Finally, we set $\beta$, $\gamma$, $\omega$ and $\mu$ to ${{10}^{-3}}$.
All images from the surround-view cameras with multiple viewpoints are shuffled thoroughly and fed to the distance and pose networks along with their respective intrinsics to create the camera geometry tensor $C_t$, as shown in Fig.~\ref{fig:mtl_pipeline}, and described in Section~\ref{sec:camera-geometry-tensor}.\par
\section{Experimental Results}

\begin{table*}[t]
\captionsetup{singlelinecheck=false, font=footnotesize}
\centering
\caption{\textbf{Ablation study on the effect of our contributions} up to our final OmniDet model on the Fisheye Woodscape dataset \cite{yogamani2019woodscape}.}
\label{table:features}
\scalebox{0.8}{
\small
\setlength{\tabcolsep}{0.3em}
\begin{tabular}{lccccccccccccc}
\toprule
\textbf{Network} &
  \textit{\begin{tabular}[c]{@{}c@{}}Robust\\ loss\end{tabular}} &
  \textit{\begin{tabular}[c]{@{}c@{}}Feature\\ loss\end{tabular}} &
  \textit{\begin{tabular}[c]{@{}c@{}}Semantic\\ Guide Dist.\end{tabular}} &
  \textit{\begin{tabular}[c]{@{}c@{}}Semantic\\ Mask\end{tabular}} &
  \textit{\begin{tabular}[c]{@{}c@{}}Motion\\ Mask\end{tabular}} &
  \textit{\begin{tabular}[c]{@{}c@{}}Semantic\\ Guide Det.\end{tabular}} &
  \textit{CGT} &
  \textit{\begin{tabular}[c]{@{}l@{}}Cyl\\ Rect. \end{tabular}} &
  \cellcolor[HTML]{00b0f0}\textbf{\textit{RMSE} $\mathbf{\downarrow}$} &
  \cellcolor[HTML]{00b0f0}$\delta<1.25 \mathbf{\uparrow}$ &
  \multicolumn{1}{c}{\cellcolor[HTML]{00b050}\begin{tabular}[c]{@{}c@{}}\textbf{\textit{mIoU}}\\ \textbf{\textit{Seg}} \end{tabular}} &
  \multicolumn{1}{c}{\cellcolor[HTML]{ab9ac0}\begin{tabular}[c]{@{}c@{}}\textbf{\textit{mIoU}}\\ \textbf{\textit{Mot.}}\end{tabular}} &
  \multicolumn{1}{c}{\cellcolor[HTML]{a5a5a5}\begin{tabular}[c]{@{}c@{}}\textbf{\textit{mAP}}\\ \textbf{\textit{Det}}\end{tabular}} \\
\midrule
\multirow{10}{*}{\begin{tabular}[c]{@{}c@{}} OmniDet \\ (SAN10-patch) \end{tabular}}
& \ch & \xm & \xm & \xm & \xm & \xm & \xm & \xm & 2.153 & 0.875 & 73.2 & 71.8 & 63.3 \\  
& \ch & \ch & \xm & \xm & \xm & \xm & \xm & \xm & 1.764 & 0.897 & 73.6 & 72.3 & 63.5 \\  
& \ch & \ch & \xm & \xm & \xm & \xm & \ch & \xm & 1.681 & 0.902 & 74.2 & 73.5 & 63.8 \\  
& \ch & \ch & \ch & \ch & \xm & \xm & \xm & \xm & 1.512 & 0.905 & 74.5 & 73.4 & 63.6 \\  
& \ch & \ch & \ch & \ch & \xm & \xm & \ch & \xm & 1.442 & 0.908 & 74.8 & 74.0 & 64.1 \\   
& \ch & \ch & \ch & \xm & \ch & \xm & \xm & \xm & 1.397 & 0.915 & 75.2 & 74.3 & 64.0 \\
& \ch & \ch & \ch & \xm & \ch & \xm & \ch & \xm & 1.352 & 0.916 & 75.5 & 74.8 & 64.3 \\ 
& \ch & \ch & \ch & \xm & \ch & \ch & \xm & \xm & 1.348 & 0.915 & 75.9 & 74.9 & 67.8 \\   
& \ch & \ch & \ch & \xm & \ch & \ch & \ch & \xm & \textbf{1.332} & \textbf{0.918} & \textbf{76.6} & \textbf{75.3} & \textbf{68.4} \\
& \ch & \ch & \ch & \xm & \ch & \ch & \ch & \ch & 1.210 & 0.929 & 78.9 & 79.2 & 74.1 \\  
\midrule
\multirow{3}{*}{\begin{tabular}[c]{@{}c@{}} OmniDet \\ (SAN10-pair) \end{tabular}}
& \ch & \ch & \ch & \ch & \xm & \xm & \ch & \xm & 1.492 & 0.904 & 74.1 & 73.1 & 63.3 \\
& \ch & \ch & \ch & \xm & \ch & \ch & \ch & \xm & 1.321 & 0.911 & 75.4 & 74.6 & 67.6 \\   
& \ch & \ch & \ch & \xm & \ch & \ch & \ch & \ch & 1.272 & 0.919 & 77.1 & 77.4 & 72.6 \\
\midrule
\multirow{10}{*}{\begin{tabular}[c]{@{}c@{}} OmniDet \\ (SAN19-patch) \end{tabular}}
& \ch & \xm & \xm & \xm & \xm & \xm & \xm & \xm & 2.138 & 0.880 & 73.9 & 72.4 & 64.7 \\  
& \ch & \ch & \xm & \xm & \xm & \xm & \xm & \xm & 1.749 & 0.903 & 74.3 & 73.0 & 64.8 \\  
& \ch & \ch & \xm & \xm & \xm & \xm & \ch & \xm & 1.662 & 0.906 & 74.6 & 74.1 & 65.2 \\  
& \ch & \ch & \ch & \ch & \xm & \xm & \xm & \xm & 1.495 & 0.910 & 74.9 & 73.8 & 64.9 \\  
& \ch & \ch & \ch & \ch & \xm & \xm & \ch & \xm & 1.427 & 0.916 & 75.4 & 74.7 & 65.5 \\   
& \ch & \ch & \ch & \xm & \ch & \xm & \xm & \xm & 1.378 & 0.918 & 75.7 & 75.1 & 65.3 \\
& \ch & \ch & \ch & \xm & \ch & \xm & \ch & \xm & 1.331 & 0.922 & 76.2 & 75.6 & 65.9 \\ 
& \ch & \ch & \ch & \xm & \ch & \ch & \xm & \xm & 1.320 & 0.927 & 76.8 & 76.2 & 69.6 \\   
& \ch & \ch & \ch & \xm & \ch & \ch & \ch & \xm & \textbf{1.304} & \textbf{0.931} & \textbf{77.4} & \textbf{77.0} & \textbf{71.5} \\
& \ch & \ch & \ch & \xm & \ch & \ch & \ch & \ch & 1.177 & 0.938 & 80.2 & 80.5 & 76.3 \\
\bottomrule
\end{tabular}
}
\end{table*}

\subsection{Datasets}

We systematically train and test all our single and multi-task models on Woodscape~\cite{yogamani2019woodscape}, a Surround View fisheye dataset with annotations for multiple tasks and on the pinhole camera datasets KITTI~\cite{kitti} and Cityscapes~\cite{cityscapes}.\par
\subsubsection{WoodScape}

The WoodScape dataset consists of 10,000 images split into training, validation, and test in a 6:1:3 ratio. Additional proprietary data was used for pre-training and initialization of our model. We train our 2D box detection on 5 most essential categories of objects — \textit{pedestrians, vehicles, riders, traffic sign, and traffic lights}. Polygon prediction task on raw fisheye is limited to only 2 classes — \textit{pedestrians, and vehicles}. Unlike traffic lights and traffic signs, these categories are non-rigid in nature and quite diverse in appearance, making them suitable for polygon regression. We sample 24 points with high curvature values from each object instance contour for the polygon regression task. Learning these points helps to regress better polygon shapes, as these points at high curvature define the shape of the object contours. 
We perform semantic segmentation on \textit{road, lanes, and curbs categories}. Images are resized to $544 \times 288\,\mathrm{px}$ from the native $1\mathrm{MP}$ resolution.\par
\subsubsection{Cityscapes} 

In the case of Cityscapes, we extracted the 2D boxes from the instance polygons. We train our OmniDet MTL model on 2,975 images in both single-task and multi-task settings. We report all our results on the validation split consisting of 500 images. Images are resized to $640 \times 384\,\mathrm{px}$ for training and validation.\par
\subsubsection{KITTI}

The KITTI dataset consists of 42,382 stereo sequences with corresponding raw LiDAR scans, 7,481 images with bounding box annotations, and 200 training images with semantic annotations. We use the data split according to Eigen~\etal~\cite{eigen2015predicting} for self-supervised depth estimation with an input size of $640 \times 192\,\mathrm{px}$ for all the tasks. For further details on the split, refer to~\cite{kumar2020fisheyedistancenet}. For the motion segmentation, we use the annotations provided by DeepMotion~\cite{city_kitti_motion} for Cityscapes and KITTI MoSeg~\cite{siam2018modnet}. Here the labels are available only for the cars category.\par

\begin{table}[t]
\captionsetup{singlelinecheck=false, skip=1pt, font=footnotesize, belowskip=0pt}
\centering
\caption{\textbf{Comparative study of SAN10-Patch MTL model and the equivalent single task models on three datasets.}}
\label{table:mtl_ablation}
\begin{adjustbox}{width=\columnwidth}
\small
\begin{tabular}{cccccccccccc}
\toprule
\multicolumn{1}{c}{\textit{\cellcolor[HTML]{00b0f0}\begin{tabular}[c]{@{}c@{}}Dist. \& \\ Pose Est.\end{tabular}}} &
\multicolumn{1}{c}{\textit{\cellcolor[HTML]{00b050}\begin{tabular}[c]{@{}c@{}}Sem.\\ Seg.\end{tabular}}} &
\multicolumn{1}{c}{\textit{\cellcolor[HTML]{ab9ac0}\begin{tabular}[c]{@{}c@{}}Mot.\\ Seg.\end{tabular}}} &
\multicolumn{1}{c}{\cellcolor[HTML]{a5a5a5}\textit{\begin{tabular}[c]{@{}c@{}}Obj.\\ Det.\end{tabular}}} &  
\multicolumn{1}{c}{\cellcolor[HTML]{e5b9b5}\textit{\begin{tabular}[c]{@{}c@{}}Soil.\\ Seg.\end{tabular}}} &
\multicolumn{1}{c}{\cellcolor[HTML]{00b0f0}\textbf{\textit{RMSE}}} &
\multicolumn{1}{c}{\cellcolor[HTML]{00b050}\begin{tabular}[c]{@{}c@{}}\textbf{\textit{mIoU}} \\ \textbf{\textit{Seg.}}\end{tabular}} &
\multicolumn{1}{c}{\cellcolor[HTML]{ab9ac0}\begin{tabular}[c]{@{}c@{}}\textbf{\textit{mIoU}} \\ \textbf{\textit{Mot.}}\end{tabular}} &
\multicolumn{1}{c}{\cellcolor[HTML]{a5a5a5}\begin{tabular}[c]{@{}c@{}}\textbf{\textit{mAP}} \\ \textbf{\textit{Det.}}\end{tabular}} &
\multicolumn{1}{c}{\cellcolor[HTML]{e5b9b5}\begin{tabular}[c]{@{}c@{}}\textbf{\textit{mIoU}} \\ \textbf{\textit{Soil.}}\end{tabular}} &
\multicolumn{1}{c}{\textit{\cellcolor[HTML]{80CBC4}\begin{tabular}[c]{@{}c@{}}Infer.\\ (fps)\end{tabular}}} \\
\midrule
\multicolumn{11}{c}{\cellcolor[HTML]{34FF34}\textit{WoodScape}} \\
\midrule
    \ch & \xm & \xm & \xm & \xm & 1.681 & \xmb & \xmb & \xmb & \xmb & 210 \\
    \xm & \ch & \xm & \xm & \xm & \xmb  & 72.5 & \xmb & \xmb & \xmb & 190 \\
    \xm & \xm & \ch & \xm & \xm & \xmb  & \xmb & 68.1 & \xmb & \xmb & 105 \\
    \xm & \xm & \xm & \ch & \xm & \xmb  & \xmb & \xmb & 63.5 & \xmb & 190 \\
    \xm & \xm & \xm & \xm & \ch & \xmb  & \xmb & \xmb & \xmb & 80.8 & 190 \\
    \ch & \ch & \xm & \xm & \xm & 1.442 & 74.8 & \xmb & \xmb & \xmb & 143 \\
    \xm & \ch & \xm & \ch & \xm & \xmb  & 77.1 & \xmb & 67.9 & \xmb & 143 \\
    \ch & \ch & \ch & \xm & \xm & 1.352 & 75.5 & 74.8 & \xmb & \xmb & 69  \\
    \ch & \ch & \ch & \ch & \xm & \textbf{1.332} & \textbf{76.6} & \textbf{75.3} & \textbf{68.4} & \xmb & \textbf{60}  \\
\midrule
\multicolumn{11}{c}{\cellcolor[HTML]{FD6864}\textit{KITTI}} \\
\midrule
    \ch & \xm & \xm & \xm & \nap & 4.126 & \xmb & \xmb & \xmb & \nap & 160 \\
    \xm & \ch & \xm & \xm & \nap & \xmb  & 67.7 & \xmb & \xmb & \nap & 148 \\
    \xm & \xm & \ch & \xm & \nap & \xmb  & \xmb & 68.3 & \xmb & \nap & 78 \\
    \xm & \xm & \xm & \ch & \nap & \xmb  & \xmb & \xmb & 80.1 & \nap & 182 \\
    \ch & \ch & \xm & \xm & \nap & 3.984 & 72.1 & \xmb & \xmb & \nap & 103 \\
    \ch & \ch & \ch & \xm & \nap & 3.892 & 71.9 & 71.7 & \xmb & \nap & 47 \\
    \ch & \ch & \ch & \ch & \nap & \textbf{3.859} & \textbf{72.4} & \textbf{72.2} & \textbf{82.3} & \nap & \textbf{43} \\
    \midrule
\multicolumn{11}{c}{\cellcolor[HTML]{448BE9}\textit{CityScapes}}  \\
\midrule
    \ch & \xm & \xm & \xm & \nap & 4.906 & \xmb & \xmb & \xmb & \nap & 156 \\
    \xm & \ch & \xm & \xm & \nap & \xmb  & 78.7 & \xmb & \xmb & \nap & 132 \\
    \xm & \xm & \ch & \xm & \nap & \xmb  & \xmb & 70.4 & \xmb & \nap & 64 \\
    \xm & \xm & \xm & \ch & \nap & \xmb  & \xmb & \xmb & 51.7 & \nap & 167 \\
    \ch & \ch & \xm & \xm & \nap & 4.741 & 79.4 & \xmb & \xmb & \nap & 91 \\
    \ch & \ch & \ch & \xm & \nap & 4.725 & 79.1 & 72.0 & \xmb & \nap & 36 \\
    \ch & \ch & \ch & \ch & \nap & \textbf{4.691} & \textbf{81.2} & \textbf{72.7} & \textbf{53.0} & \nap & \textbf{31} \\
    \bottomrule
\end{tabular}
\end{adjustbox}
\end{table}
\subsection{Single Task vs Multi-Task Learning} \label{sec:single_vs_mtl}

In Table~\ref{table:mtl_ablation}, we perform an extensive ablation of our proposed framework on all the datasets, as mentioned above. Quantitative results from our experiments indicate that a multi-task network with 6 tasks, 5 diverse tasks perform better than the single task models along with our proposed synergies explained in Section~\ref{sec:synergy}. For KITTI and CityScapes, we employ our novel VarNorm task weighting technique. With this synergy of perception tasks, we obtain state-of-the-art depth and pose estimation results on the KITTI dataset, as shown in Table~\ref{tab:kitti-results} and Table~\ref{table:pose-ate} respectively. We infer our models using the TensorRT (FP16bit) on NVIDIA's Jetson AGX platform and report the FPS for all the tasks.\par

\subsection{Ablation Study of our Contributions}
\label{sec:features-ablation}

For our ablation analysis of the main features shown in Table~\ref{table:features}, we consider two variants of the self-attention encoder, namely pairwise and patchwise. First, we replace the \lone loss with a generic parameterized loss function and test it using the self-attention encoder's patchwise variant. We cap the distance estimates to 40m. We achieve significant gains in this setting by attributing better-supervised signal provided by using discriminative features loss $\mathcal{L}_{dis}$. In this case, incorrect distance values are appropriately penalized with more considerable losses along with the combination of $\mathcal{L}_{cvt}$ wherein a correct optimization direction is provided. These losses help the gradient descent approaches to transit smoothly towards optimal solutions. When adding the camera geometry tensor to this setting, we observe a significant increase in accuracy since we train multiple cameras with different camera intrinsics and viewing angles. For the OmniDet framework to be operational in the first place, this an important feature. The aforementioned training strategy makes the network camera-independent and better generalizes to images taken from a different camera.\par

\begin{table}[t]
\captionsetup{singlelinecheck=false, font=footnotesize}
\centering{
\caption{\textbf{Comparison of task-weighting methods on the WoodScape dataset. PA denotes pixel accuracy.}}
\label{table:task-weighting}
\begin{adjustbox}{width=\columnwidth}
\small
\setlength{\tabcolsep}{0.15em}
\begin{tabular}{lccccccc}
\toprule
  \multicolumn{1}{c|}{\textit{\begin{tabular}[c]{@{}c@{}}Task\\ Weighting\end{tabular}}} &
  \multicolumn{2}{c|}{\cellcolor[HTML]{00b0f0}\textit{\begin{tabular}[c]{@{}c@{}}Distance \\ Estimation\end{tabular}}} &
  \multicolumn{2}{c|}{\cellcolor[HTML]{00b050}\textit{\begin{tabular}[c]{@{}c@{}}Semantic \\ Segmentation\end{tabular}}} &
  \multicolumn{2}{c|}{\textit{\cellcolor[HTML]{ab9ac0}\begin{tabular}[c]{@{}c@{}}Motion\\ Segmentation\end{tabular}}} &
  \multicolumn{1}{c|}{\cellcolor[HTML]{a5a5a5}\textit{\begin{tabular}[c]{@{}c@{}}Object\\ Detection\end{tabular}}} \\ 
\midrule
  &
  \multicolumn{1}{|c|}{\textbf{Sq. Rel $\mathbf{\downarrow}$}} &
  \multicolumn{1}{c|}{\textbf{Abs Rel $\mathbf{\downarrow}$}} &
  \multicolumn{1}{c|}{\textbf{mIoU $\mathbf{\uparrow}$}} &
  \multicolumn{1}{c|}{\textbf{PA $\mathbf{\uparrow}$}} &
  \multicolumn{1}{c|}{\textbf{mIoU} $\mathbf{\uparrow}$} &
  \multicolumn{1}{c|}{\textbf{PA} $\mathbf{\uparrow}$} &
  \multicolumn{1}{c|}{\textbf{mAP} $\mathbf{\uparrow}$} \\
\midrule

Single Task                               & 0.060  & 0.304 & 72.5 & 94.8 & 68.1 & 94.1 & 63.5 \\
\midrule
Equal                                     & 0.058  & 0.302 & 70.3 & 92.7 & 67.3 & 93.3 & 64.6 \\
DTP~\cite{guo2018dynamic}                 & 0.047  & 0.281 & 75.8 & 95.6 & 75.3 & 95.3 & 67.9 \\
DWA~\cite{liu2019end}                     & 0.054  & 0.293 & 75.4 & 95.2 & 74.7 & 95.1 & 67.5 \\
Geometric~\cite{chennupati2019multinet++} & 0.061  & 0.297 & 74.2 & 94.1 & 73.2 & 94.3 & 66.7 \\
GradNorm~\cite{chen2018gradnorm}          & 0.050  & 0.283 & 75.9 & 95.7 & 74.9 & 96.0 & 67.7 \\
Uncertainity~\cite{Kendall2018}           & 0.044 & 0.279  & 76.1 & 96.2 & 75.1 & 95.8 & 68.0 \\
\textbf{VarNorm}                          & \textbf{0.046} & \textbf{0.276}  & \textbf{76.6} & \textbf{96.4} & \textbf{75.3} & \textbf{96.1} & \textbf{68.4} \\ 
\bottomrule
\end{tabular}
\end{adjustbox}
}
\end{table}

To achieve synergy between geometry and semantic features, we add semantic guidance to the distance decoder. It helps to reason about geometry and content within the same shared features and disambiguate photometric ambiguities. To establish a robust reconstruction loss free from the dynamic objects' contamination, we introduce semantic and motion masks as described in Section~\ref{sec:dynamic-object-mask}, to filter all the dynamic objects. Motion mask based filtering yields superior gains along with CGT compared to using semantic masks as semantics might not contain all the dynamic objects in its set of classes as indicated in Eq.~\ref{eq:semantic_mask}. Additionally, this contribution possesses the potential to solve the infinite distance issue. Finally, to complete our synergy, we use semantically guided features to the detection decoder as described in Section~\ref{sec:semantic-guidance}, which yields significant gains in mAP, and overall results for all the tasks are inherently improved with better-shared features. All our contributed features and the synergy between tasks help the OmniDet framework to achieve a good scene understanding with high accuracy in each task's predictions. We also experiment using cylindrical rectification (Cyl Rect.), which provides a good trade-off between loss of field-of-view and reducing distortion \cite{yogamani2019woodscape}.\par

For object detection on native fisheye images, in addition to the standard 2D box representation, we benchmark oriented boxes, ellipse, curved boxes, and 24-sided polar polygon representations in Table~\ref{tab:fisheye-yolov3}. Here mIoU GT represents the maximum performance we can achieve in terms of instance segmentation by using each representation. It is computed between the ground truth instance segmentation and the ground truth of the corresponding representation. Whereas mIoU represents the performance achieved with our network. We also list the number of parameters involved for each representation for the complexity comparison.\par
\begin{table}[!t]
\captionsetup{singlelinecheck=false, font=footnotesize, skip=2pt}
\centering
\caption{\textbf{Evaluation of various object detection representations.}}
\scalebox{0.65}{
\begin{adjustbox}{width=\columnwidth}
\small
\begin{tabular}{lccc}
\toprule
\textit{\textbf{Representation}} 
& \multicolumn{1}{c}{\cellcolor[HTML]{7d9ebf}\begin{tabular}[c]{@{}c@{}}\textit{mIoU}\\\textit{GT}\end{tabular}} & 
\textit{\cellcolor[HTML]{7d9ebf}mIOU} & 
\cellcolor[HTML]{7d9ebf}\begin{tabular}[c]{@{}l@{}}\textit{No. of}\\ \textit{params}\end{tabular} \\ 
\midrule
Standard Box     & 51.3 & 31.6 & 4 \\
Curved Box       & 52.5 & 32.3 & 6 \\
Oriented Box     & 53.9 & 33.6 & 5 \\
Ellipse          & 55.5 & 35.4 & 5 \\
\textbf{24-sided Polygon} & \textbf{86.6} & \textbf{44.6} & \textbf{48} \\
\bottomrule
\end{tabular}
\end{adjustbox}
}
\label{tab:fisheye-yolov3}
\vspace{-1em}
\end{table} 
\subsection{State-of-the-Art Comparison on KITTI} \label{sec:kitti-depth-pose}

To facilitate comparison to previous methods, we also train our distance estimation method in the classical depth estimation setting on the KITTI Eigen split~\cite{geiger2013vision} whose results are shown in Table~\ref{tab:kitti-results}. With the synergy between depth, semantic, motion, and detection tasks along with the features ablated in Table~\ref{table:features} and their importance explained in the Section~\ref{sec:features-ablation}, \textit{we outperform all the previous monocular methods.}
Following best practices, we cap depths at 80\,m. We also evaluate using the \textit{Original}~\cite{Eigen_14} as well as \textit{Improved}~\cite{uhrig2017sparsity} ground truth depth maps. Method${^*}$ indicates the online refinement technique~\cite{Casser2019}, where the model is trained during the inference. Using the online refinement method from~\cite{Casser2019}, we obtain a significant improvement.\par

In Table~\ref{table:pose-ate}, we report the average trajectory error in meters of the pose estimation network by following the same protocols by Zhou~\cite{zhou2017unsupervised} on the official KITTI odometry split (containing 11 sequences with ground-truth (GT) odometry acquired with the IMU/GPS measurements, which is used for evaluation purpose only), and use sequences 00-08 for training and 09-10 for testing. We outperform the previous methods listed in Table~\ref{table:pose-ate}, mainly by applying our bundle adjustment framework using our cross-sequence distance consistency loss~\cite{kumar2020fisheyedistancenet} that induces more constraints and simultaneously optimizes distances and camera pose for an implicitly extended training input sequence. This provides additional consistency constraints that are not induced by previous methods.\par
\begin{table}[!t]
\captionsetup{singlelinecheck=false, font=footnotesize}
\centering
\caption{\textbf{Evaluation of depth estimation on the KITTI Eigen split}.}
\label{tab:kitti-results}
\begin{adjustbox}{width=\columnwidth}
\small
\begin{tabular}{c|lcccccccc}
\toprule
  & 
  \multicolumn{1}{c}{\textbf{Method}} &
  \multicolumn{1}{c}{\textit{\cellcolor[HTML]{7d9ebf}Abs$_{rel}$}} &
  \textit{\cellcolor[HTML]{7d9ebf}Sq$_{rel}$} &
  \multicolumn{1}{c}{\textit{\cellcolor[HTML]{7d9ebf}RMSE}} &
  \multicolumn{1}{c}{\textit{\cellcolor[HTML]{7d9ebf}\begin{tabular}[c]{@{}c@{}}RMSE$_{log}$\end{tabular}}} &
  \multicolumn{1}{c}{\textit{\cellcolor[HTML]{e8715b}$\delta_1$}} &
  \multicolumn{1}{c}{\textit{\cellcolor[HTML]{e8715b}$\delta_2$}} &
  \multicolumn{1}{c}{\textit{\cellcolor[HTML]{e8715b}$\delta_3$}} \\ 
  \cmidrule(l){3-6} \cmidrule(lr){7-9}
  & & \multicolumn{4}{c}{\cellcolor[HTML]{7d9ebf}lower is better} & \multicolumn{3}{c}{\cellcolor[HTML]{e8715b}higher is better} \\
\midrule
\parbox[t]{2mm}{\multirow{10}{*}{\rotatebox[origin=c]{90}{Original~\cite{Eigen_14}}}}
& Monodepth2~\cite{Godard2019}               & 0.115 & 0.903 & 4.863 & 0.193 & 0.877 & 0.959 & 0.981 \\
& PackNet-SfM~\cite{Guizilini2020a}          & 0.111 & 0.829 & 4.788 & 0.199 & 0.864 & 0.954 & 0.980 \\
& FisheyeDistanceNet~\cite{kumar2020fisheyedistancenet} & 0.117 & 0.867 & 4.739 & 0.190 & 0.869 & 0.960 & 0.982 \\
& UnRectDepthNet~\cite{kumar2020unrectdepthnet} & 0.107 & 0.721 & 4.564 & 0.178 & 0.894 & 0.971 & \textbf{0.986} \\
& SynDistNet~\cite{kumar2020syndistnet}      & 0.109 & 0.718 & 4.516 & 0.180 & 0.896 & 0.973 & \textbf{0.986} \\
& Shu \etal~\cite{shu2020featdepth}          & 0.104 & 0.729 & 4.481 & 0.179 & 0.893 & 0.965 & 0.984 \\
& OmniDet                                    & \textbf{0.092} & \textbf{0.657} & \textbf{3.984} & \textbf{0.168} & \textbf{0.914} & \textbf{0.975} & \textbf{0.986} \\ 
\cmidrule{2-9}
& Struct2Depth${^*}$~\cite{Casser2019}       & 0.109 & 0.825 & 4.750 & 0.187 &0.874 & 0.958 & 0.983 \\
& GLNet${^*}$~\cite{Chen2019b}               & 0.099 & 0.796 & 4.743 & 0.186 &0.884 & 0.955 & 0.979 \\
& Shu${^*}$ \etal~\cite{shu2020featdepth}    & 0.088 & 0.712 & 4.137 & 0.169 & 0.915 & 0.965 & 0.982 \\
& OmniDet${^*}$                              & \textbf{0.077} & \textbf{0.641} & \textbf{3.859} & \textbf{0.152} & \textbf{0.931} & \textbf{0.979} & \textbf{0.989} \\
\midrule
\parbox[t]{2mm}{\multirow{6}{*}{\rotatebox[origin=c]{90}{Improved~\cite{uhrig2017sparsity}}}}
& Monodepth2~\cite{Godard2019}               & 0.090 & 0.545 & 3.942 & 0.137 & 0.914 & 0.983 & 0.995 \\
& PackNet-SfM~\cite{Guizilini2020a}          & 0.078 & 0.420 & 3.485 & 0.121 & 0.931 & 0.986 & 0.996 \\
& UnRectDepthNet~\cite{kumar2020unrectdepthnet} & 0.081 & 0.414 & 3.412 & 0.117 & 0.926 & 0.987 & 0.996 \\
& SynDistNet~\cite{kumar2020syndistnet}      & 0.076 & 0.412 & 3.406 & 0.115 & 0.931 & 0.988 & 0.996 \\
& OmniDet                                    & \textbf{0.067} & \textbf{0.306} & \textbf{3.098} & \textbf{0.101} & \textbf{0.944} & \textbf{0.991} & \textbf{0.997} \\
& OmniDet${^*}$                              & \textbf{0.048} & \textbf{0.287} & \textbf{2.913} & \textbf{0.081} & \textbf{0.948} & \textbf{0.991} & \textbf{0.998} \\
\bottomrule
\end{tabular}
\end{adjustbox}
\end{table}
\begin{table}[t]
\captionsetup{singlelinecheck=false, font=footnotesize, skip=2pt, belowskip=0pt}
\centering
\caption{\textbf{Evaluation of the pose estimation} on the KITTI Odometry Benchmark~\cite{geiger2013vision}.}
\scalebox{0.95}{
\begin{adjustbox}{width=\columnwidth}
\begin{tabular}{lcccc}
\toprule
\multicolumn{1}{l}{\textbf{Method}} 
& \textit{\begin{tabular}[c]{@{}c@{}}No. of\\ Frames\end{tabular}} 
& \textit{GT} 
& \cellcolor[HTML]{7d9ebf}\textit{Sequence 09} 
& \cellcolor[HTML]{e8715b}\textit{Sequence 10} \\ 
\midrule
    GeoNet~\cite{Yin2018}                        & 5 & \ch & 0.012 $\pm$ 0.007 & 0.012 $\pm$ 0.009 \\
    Struct2Depth~\cite{Casser2019}               & 5 & \ch & 0.011 $\pm$ 0.006 & 0.011 $\pm$ 0.010 \\
    Ranjan~\cite{Ranjan2019}                     & 5 & \ch & 0.011 $\pm$ 0.006 & 0.011 $\pm$ 0.010 \\ 
    PackNet-SfM~\cite{Guizilini2020a}            & 5 & \ch & 0.010 $\pm$ 0.005 & 0.009 $\pm$ 0.008 \\
    PackNet-SfM~\cite{Guizilini2020a}            & 5 & \xm & 0.014 $\pm$ 0.007 & 0.012 $\pm$ 0.008 \\
    OmniDet                                      & 5 & \ch & \textbf{0.009} $\pm$ \textbf{0.004} & 0.008 $\pm$ \textbf{0.005} \\
    OmniDet                                      & 5 & \xm & \textbf{0.010} $\pm$ \textbf{0.005} & \textbf{0.010} $\pm$ \textbf{0.008} \\
    \bottomrule
\end{tabular}
\end{adjustbox}
}
\label{table:pose-ate}
\vspace{-1em}
\end{table}
\section{Conclusion}

We successfully demonstrated a six-task network with a shared encoder and synergized decoders on fisheye surround-view images in this work. The majority of the automated driving community's research continues to focus on individual tasks, and there is much progress to be made in designing and training optimal multi-task models. We have several novel contributions, including camera geometry tensor usage for encoding radial distortion and variance-based normalization task weighting, and generalized object detection representations. To enable comparison, we evaluate our network on five tasks on KITTI and Cityscapes, achieving competitive results. There are still many practical challenges in scaling to a higher number of tasks: building a diverse and balanced dataset, corner case mining, stable training mechanisms, and designing an optimal map representation that combines all tasks reducing the post-processing. We hope that this work encourages further research in building a unified perception model for autonomous driving.\par
\vspace{-0.6em}
\bibliographystyle{IEEEtran}
\bibliography{bib/backup}
\end{document}